\RequirePackage{fix-cm}
\documentclass[twocolumn]{svjour3}          
\smartqed  
\usepackage{graphicx}
\usepackage{cite}
\usepackage{latexsym}
\usepackage{framed,multirow}
\usepackage{subfigure}
\usepackage{dingbat}
\usepackage{xcolor}
\usepackage{multicol}
\usepackage{hyperref}
\usepackage{amsfonts}

\definecolor{mypink1}{rgb}{0.858, 0.188, 0.478}
\begin{document}

\title{Multiple Convolutional Features in Siamese Networks for Object Tracking
}


\author{Zhenxi Li         \and
        Guillaume-Alexandre Bilodeau \and
        Wassim Bouachir
}


\institute{Zhenxi Li \at
              LITIV lab, Polytechnique Montreal, Montreal, H3T 1J4, Canada \\
              \email{zhenxili96@gmail.com}           
           \and
          Guillaume-Alexandre Bilodeau \at
              LITIV lab, Polytechnique Montreal, Montreal, H3T 1J4, Canada \\
              \email{gabilodeau@polymtl.ca}
            \and
          Wassim Bouachir \at
            TELUQ University, Montreal, H2S 3L5, Canada \\
            \email{wassim.bouachir@teluq.ca}
}

\date{Received: date / Accepted: date}

\maketitle

\begin{abstract}
Siamese trackers demonstrated high performance in object tracking due to their balance between accuracy and speed. Unlike classification-based CNNs, deep similarity networks are specifically designed to address the image similarity problem, and thus are inherently more appropriate for the tracking task. However, Siamese trackers mainly use the last convolutional layers for similarity analysis and target search, which restricts their performance. In this paper, we argue that using a single convolutional layer as feature representation is not an optimal choice in a deep similarity framework. We present a Multiple Features-Siamese Tracker (MFST), a novel tracking algorithm exploiting several hierarchical feature maps for robust tracking. Since convolutional layers provide several abstraction levels in characterizing an object, fusing hierarchical features allows to obtain a richer and more efficient representation of the target. Moreover, we handle the target appearance variations by calibrating the deep features extracted from two different CNN models. Based on this advanced feature representation, our method achieves high tracking accuracy, while outperforming the standard siamese tracker on object tracking benchmarks. The source code and trained models are available at \texttt{\color{mypink1} \url{https://github.com/zhenxili96/MFST}}.
\end{abstract}

\section{Introduction}
\label{intro}
Visual object tracking (VOT) is a fundamental task in computer vision. Given a target object in the first frame, the objective of VOT is to determine the object state, typically its bounding box, in the following frames. With the rapid development of computer vision, visual object tracking has been employed in many applications, such as autonomous driving, visual analysis and video surveillance. For example, with the help of visual object tracking, autonomous driving systems can analyze obstacle movements and decide where to go.

Nowadays, most successful state-of-the-art trackers are based on correlation filters (e.g. \cite{zheng2019dynamically}), deep neural networks (e.g. \cite{wang2020tracking}), or on a combination of both techniques (e.g. \cite{zgaren2020coarse}). In this work, we are particularly interested in deep learning trackers, that achieved impressive performance while bringing new ideas to VOT. This paradigm has become successful mainly due the use of convolutional neural network (CNN)-based features for appearance modeling and their discriminative ability to represent target objects. While several tracking methods use classification-based CNN models that are built following the principals of visual classification tasks, another approach \cite{SiamFC} formulates the tracking task as a deep similarity learning problem, where a Siamese network is trained to locate the target within a search image region. This method uses feature representations extracted by CNNs and performs correlation operation with a sliding window to calculate a similarity map for finding the target location. Rather than detecting by correlation, other deep similarity trackers \cite{SiamRPN,DaSiamRPN,GOTURN, iccv19_SPLT} generate the bounding box for the target object with regression networks. For example, GOTURN \cite{GOTURN} predicts the bounding box of the target object with a simple CNN model. The trackers \cite{DaSiamRPN} and \cite{SiamRPN} generate a number of proposals for the target after extracting feature representations. Classification and regression procedures are then applied to produce the final object location. The SPLT tracker \cite{iccv19_SPLT} uses a similar approach, but includes also a re-detection module for long-term tracking.

By formulating object tracking as a deep similarity learning problem, Siamese trackers achieved significant progress in terms of both speed and accuracy. However, one weakness of the siamese trackers is that they typically use only features from the last convolutional layers for similarity analysis and target state prediction. Therefore, the object representation is not as robust as it could be to target appearance variations, and tracking can be loss in more difficult scenarios. To address this weakness, we argue that using the last convolutional layers is not the optimal choice, and we demonstrate in this work that features from earlier layers are also beneficial for more accurate tracking with Siamese trackers. 

Indeed, the combination of several convolutional layers was shown to be efficient for robust tracking \cite{HCFT,HCFTX}. As we go deeper in a CNN, the receptive field becomes wider, therefore, features from different layers contain different levels of information. In this way, the last convolutional layers retain general characteristics represented in a summarized fashion, while the first convolutional layers provide low-level features. These latter are extremely valuable for precise localization of the target as they are more object-specific and capture spatial details. Furthermore, instead of using features from a single CNN model, we propose to exploit different models within the deep similarity framework. Diversifying feature representations significantly improves tracking performance. Such strategy is shown to ensure a better robustness against target appearance variations, one of the most challenging tracking difficulties \cite{MBST}.

Based on these principles, we propose a Multiple Features-Siamese Tracker (MFST). Our tracker utilizes diverse features from several convolutional layers, two models and a proper feature fusing strategies to improve tracking performance. 

Our contributions can be summarized as follows:
\begin{itemize}
    \item We propose a new tracking method that exploits feature representations from several hierarchical convolutional layers as well as different CNN models for object tracking.
    \item We propose feature fusing strategies with a feature recalibration module to make a better use of the feature representations.
    \item We show that our two previous contributions improve tracking by testing our MFST tracking algorithm on popular OTB benchmarks. We show that our method improves over the SiamFC base model and that our method achieves strong performance with respect to recent state-of-the-art trackers on popular OTB benchmarks.
\end{itemize}

The paper is organized as follows. We present the related work in Section \ref{sec:mfst_related}, the proposed MFST tracker in Section \ref{sec:mfst_main}, and the experimental results in Section \ref{sec:mfst_exp} respectively. Finally, Section \ref{sec:mfst_conclude} concludes the paper.

\section{Related Work}\label{sec:mfst_related}

\subsection{Siamese Trackers}
VOT can be formulated as a similarity learning problem. Once the deep similarity network is trained during an offline phase to learn a general similarity function, the model is applied for online tracking by analyzing the similarity between the two network inputs: the target template and the current
frame. The pioneering work, SiamFC \cite{SiamFC}, applied two identical branches made up of fully convolutional neural networks to extract the feature representations, on which cross-correlation is computed to generate the tracking result. SiamFC outperformed most of the best trackers at that time, while achieving real-time speed. Rather than performing correlation on deep features directly, CFNet \cite{CFNet} trains a correlation filter based on the extracted features of the object to speed up tracking without accuracy drop. MBST \cite{MBST} improved the tracking performance by using multiple siamese networks as branches to enhance the diversity of the feature representation. SA-Siam \cite{SASiam} encodes the target by a semantic branch and an appearance branch to improve the robustness of tracking. However, since these siamese trackers only take the output of the last convolutional layers, more-detailed target specific information from earlier layers is not used. In contrast, in our work, we adopt a Siamese architecture to extract deep features for the target and search region, but combine features from different layers of the networks for tracking.

\begin{figure*}[!t]
    \centering
    \includegraphics[width=0.8\linewidth]{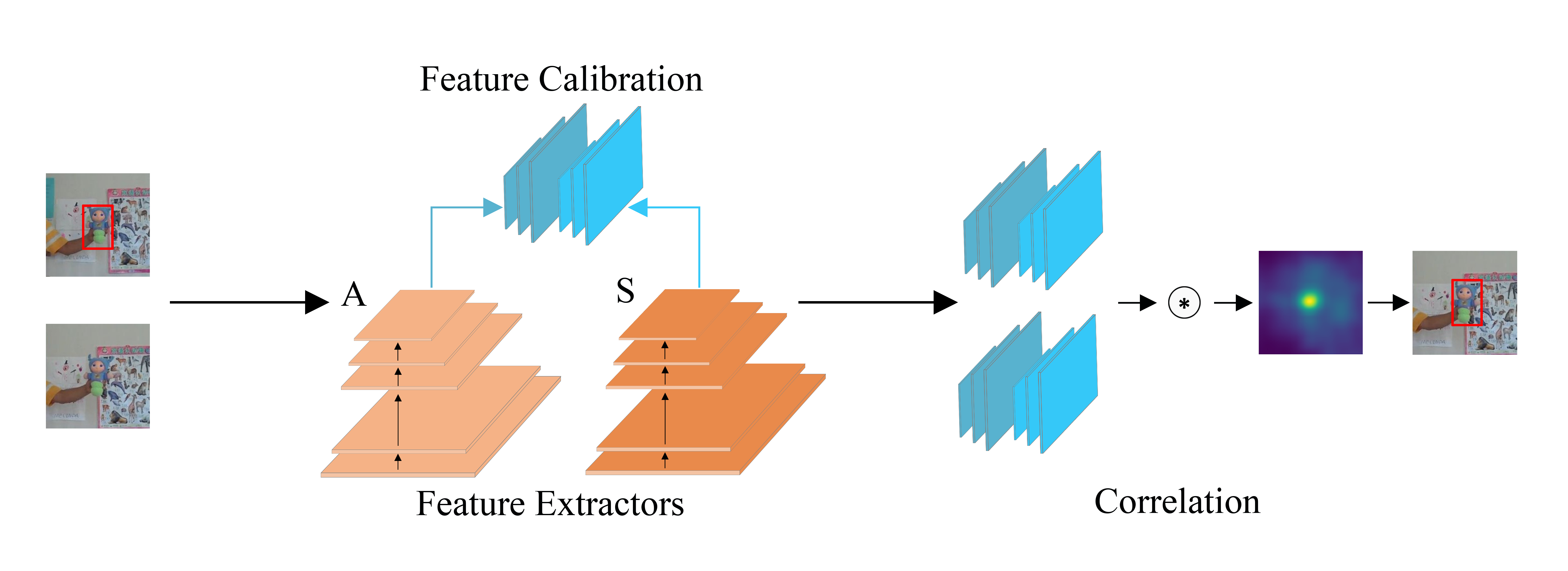}
    \caption{The architecture of our MFST tracker. Two CNN models are used as feature extractors and their features are calibrated by Squeeze-and-Excitation (SE) blocks. Then, correlations are applied over the features of the search region with the features of the exemplar patch and the output response maps are fused to calculate the new position of the target. Bright orange and blue: SiamFC (S) and dark orange and blue: AlexNet (A).}
    \label{fig:architecture}
\end{figure*}

\subsection{Hierarchical Convolutional Features in Tracking}
Most CNN-based trackers only use the output of the last convolutional layer that contains semantic information represented in a summarized fashion. However, different convolutional layers embed different levels of visual abstraction. In fact, convolutional layers provide several detail levels in characterizing an object, and the combination of different convolutional levels is demonstrated to be efficient for robust tracking \cite{HCFT, li2018deep}. In this context, the pioneering algorithm, HCFT \cite{HCFT}, tracks the target using correlation filters learned on several layers. With HCFT, the representation ability of hierarchical convolutional features is demonstrated to be better than features from a single layer. Subsequently, \cite{HCFTX} presented a visualization of features extracted from different convolutional layers. In their work, they employed three convolutional layers as the target object representations, which are then convolved with the learned correlation filters to generate the response map, and a long-term memory filter to correct results. The use of hierarchical convolutional features is shown to make their trackers much more robust. In a similar way, the SiamRPN++ \cite{Li2019SiamRPNEO} tracker uses features from several layers of a very deep network to regress the target location. The regression results obtained with several SiamRPN blocks \cite{SiamRPN}, applied each on a selected layer, are combined to obtain the final object location.

\subsection{Multi-Branch Tracking} 
One of the most challenging problem in object tracking is the varying appearance of the tracked objects. A single fixed networks cannot guarantee to generate discriminative feature representations in all tracking situations. To handle the problem of target appearance variations, TRACA \cite{TRACA} trained multiple auto-encoders, each for different appearance categories. These auto-encoders compress the feature representation for each category. The best expert auto-encoder is selected by a pretrained context-aware network. By selecting a specific auto-encoder for the tracked object, a more robust representation can be generated. MDNet \cite{MDNet} applied a fixed CNN for feature extraction, but used multiple regression branches for objects belonging to different tracking scenarios. More recently, MBST \cite{MBST} extracted the feature representation for the target object through multiple branches and selected the best branch according to their response maps. With multiple branches, MBST can obtain diverse feature representations and select the most discriminative one under the prevailing circumstance. In their study, we can observe that the greater the number of branches, the more robust the tracker is. However, this is achieved at the cost of a higher computational time. In this work, we can get a diverse feature representation of a target at lower cost because some of the representations are extracted from the many layers of the same CNN. Therefore, we do not need a large number of siamese branches.

\section{Multiple Features-Siamese Tracker}\label{sec:mfst_main}
We propose a Multiple Features-Siamese Tracker (MFST) for object tracking. For the design of our method, we considered that features from different convolutional layers contain different level of abstractions and that the different channels of the features play different roles in tracking. Furthermore, we recalibrate the deep features extracted from the CNN models and combine hierarchical features to make a more robust representation. Besides, since models trained for different tasks can diversify the feature representation as well, we build our siamese architecture with two CNN models to achieve better performance. The code of our tracker can be found at \url{https://github.com/zhenxili96/MFST}.

\subsection{Network Architecture}
As many recent object tracking approaches \cite{SiamFC, CFNet, MBST}, we formulate the tracking problem as a similarity learning problem and utilize a siamese architecture to address it. The network architecture of our tracker is shown in Figure \ref{fig:architecture}. It uses two pretrained CNN models as feature extractors, SiamFC \cite{SiamFC} and AlexNet \cite{AlexNet}, as indicated in Figure \ref{fig:architecture}. The two models are denoted as $S$ and $A$, respectively, in the following. Both of them are five layers fully convolutional neural networks.

\begin{figure*}[t]
    \centering
    \includegraphics[width=0.5\linewidth]{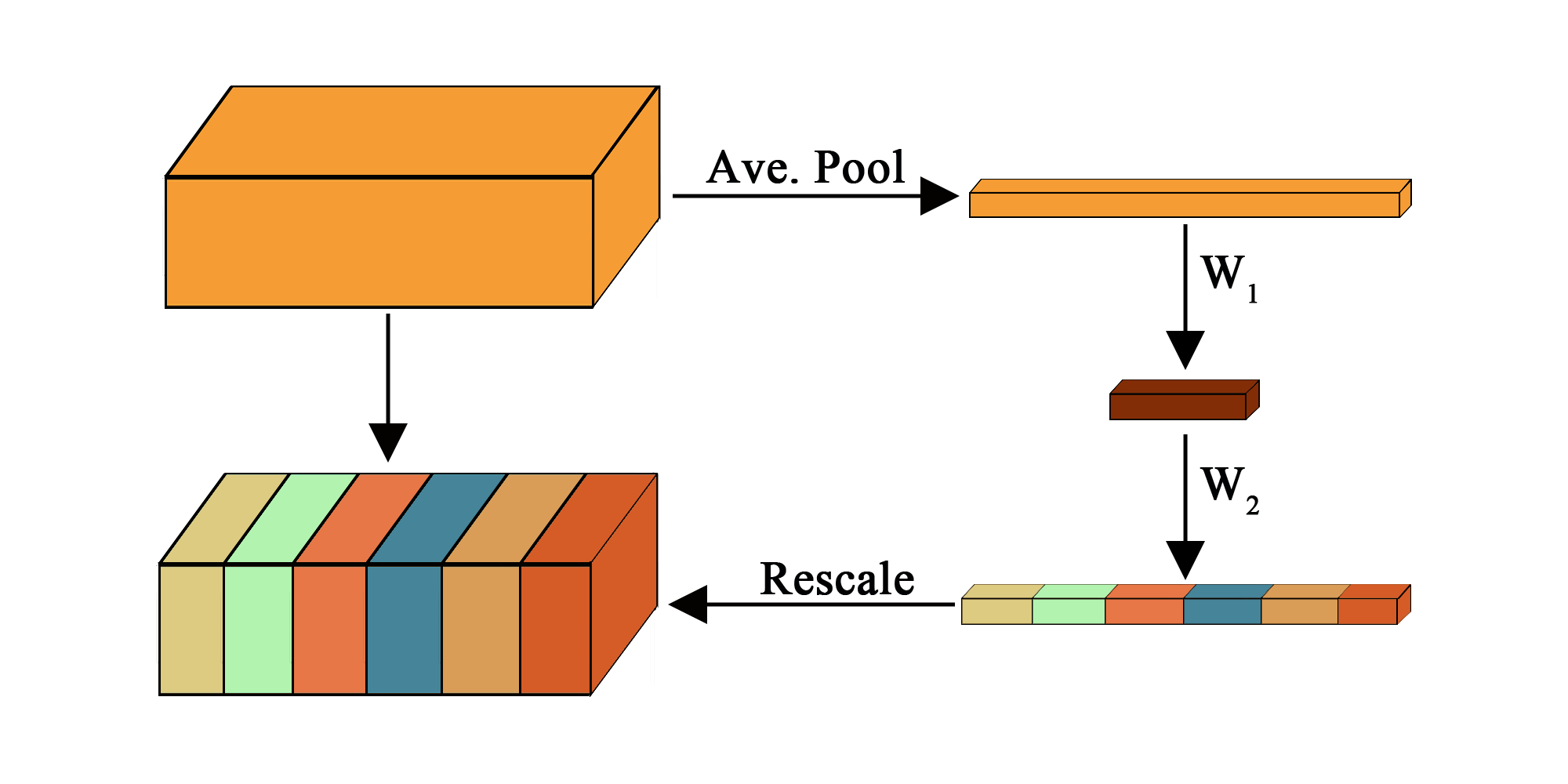}
    \caption{Illustration of a SE-block. It consists of two step, squeeze step and excitation step. The squeeze step uses average pooling operation to generate the channel descriptor, and the excitation step uses a two layers MLP to capture the channel-wise dependencies.}
    \label{fig:seblock}
\end{figure*}

The input of our method consists of an exemplar patch $z$ cropped according to the initial bounding box or the result of last frame and search region $x$. The exemplar patch has a size of $W_{z}\times H_{z}\times 3$ and the search region has a size of $W_{x}\times H_{x}\times 3$ ($W_{z}<W_{x}$ and $H_{z}<H_{x}$), representing the width, height and the color channels of the image patches. 

With the two CNN models, we obtain the deep features $S_{l_{i}}$, $A_{l_{i}}$ ($l=c3,c4,c5$, $i=z,x$) from the conv3, conv4 and conv5 layers of each model. These are the preliminary deep feature representations of the inputs. Then, these features are recalibrated through Squeeze-and-Excitation blocks (SE-blocks) \cite{SENet}. The recalibrated features are denoted as $S_{l_{i}}^{*}$, $A_{l_{i}}^{*}$, respectively, for the two models. The detail of a SE-block is illustrated in Fig. \ref{fig:seblock}. These blocks are trained to explore the importance of the different channels for tracking. They learn weights for the different channels to recalibrate features extracted from the preliminary feature representations. 

Once the recalibrated feature representations $S_{l_{i}}^{*}$ and $A_{l_{i}}^{*}$ are generated, we apply cross-correlation operations for each recalibrated feature map pairs to generate response maps. The cross-correlation operation can be implemented by a convolution layer using the features of the exemplar as a filter. Then we fuse these response maps to produce the final response map. The corresponding location of the maximum value in the response map is the new position of the target object.

Similarly to \cite{SiamFC}, the SiamFC feature extractor, as well as the SE-blocks for both feature extractors are trained with a logistic loss. For a pair of patches $z,x$, the total loss for a response map $r$ is 
\begin{equation}
    L(y,v) =\frac{1}{|r|}\sum_{u\in r} l(y[u].v[u]),
    \label{mainloss}
\end{equation}
with 
\begin{equation}
l(y,v)=log(1+exp(-yv)),
\end{equation}
where $y$ is a ground-truth label (1 or -1, for positive and negative pairs) and $v$ is a cross-correlation score at coordinate $u$ in response map $r$.
\subsection{Feature Extraction}

\paragraph{Hierarchical Convolutional Features.} It is well known that the last convolutional layer encodes more semantic information that is invariant to significant appearance variations, compared to earlier layers. However, its resolution is coarse due to the large receptive field, and it is not the most appropriate for precise localization as required in tracking. On the contrary, features from earlier layers contain less semantic information, but they retain more spatial details and they are more precise in localization. Thus, we propose to exploit multiple hierarchical levels of features to build a better representation of the target.

In our work, we use the convolutional layers of two CNN models as feature extractors, that is SiamFC \cite{SiamFC} and AlexNet \cite{AlexNet}. Each model is trained for a different task, object tracking for SiamFC and image classification for AlexNet. 
We take features extracted from the 3rd, 4th, 5th layers as the preliminary target representations.

\begin{table*}[!t]
    \centering
    \caption{Experiments with several variations of our method, where A and S denote using AlexNet or using SiamFC as the base feature extractor. \textbf{Boldface} indicates best results.}
    \begin{tabular}{|c|c c c c c|c c|c c|c c|}
    \hline
       & \multicolumn{3}{c}{Layers} & & &
      \multicolumn{2}{c|}{OTB-2013} &
      \multicolumn{2}{c|}{OTB-50} &
      \multicolumn{2}{c|}{OTB-100}\\
        Model& c3 & c4 & c5 & Fusion & SE &  AUC & Prec. & AUC & Prec. & AUC & Prec. \\
        \hline
        A&\checkmark & & & & & 0.587&0.740 & 0.474& 0.618& 0.559& 0.712\\
        A&\checkmark & & & & \checkmark & 0.603 & 0.755 & 0.504 & 0.642 & 0.587 & 0.747\\
        A& & \checkmark & & & & 0.632 & 0.789& 0.536 & 0.692 & 0.614 & 0.778 \\
        A& & \checkmark & & & \checkmark & \textbf{0.637} & 0.801 & 0.544 & 0.707 & 0.623 & 0.795 \\
        A& & & \checkmark & & & 0.582&0.763 & 0.496& 0.665&0.557 & 0.735\\
        A& & & \checkmark & &\checkmark& 0.573 & 0.762 & 0.507 & 0.696 & 0.575 & 0.769\\
        A& \checkmark & \checkmark & \checkmark & HW & & 0.623 & 0.774 & 0.515 & 0.657 & 0.605 & 0.763 \\
        A& \checkmark & \checkmark & \checkmark & SM & & 0.633 & 0.797& 0.542 & 0.705 & 0.616 & 0.784\\
        A& \checkmark & \checkmark & \checkmark & SW & & 0.630 & 0.795 & 0.538 & 0.699 & 0.616 & 0.786 \\
        A& \checkmark & \checkmark & \checkmark & HW & \checkmark & 0.627 & 0.798 & 0.537 & 0.700 & 0.617 & 0.790 \\
        A& \checkmark & \checkmark & \checkmark & SM & \checkmark & 0.631 & 0.799 & 0.542 & 0.706 & 0.621 & 0.792 \\
        A& \checkmark & \checkmark & \checkmark & SW & \checkmark & 0.635 & \textbf{0.811} & \textbf{0.545} & \textbf{0.716} & \textbf{0.627} & \textbf{0.803} \\
         \hline
        \hline
        S& \checkmark & & & & & 0.510 & 0.661 & 0.439 & 0.574 & 0.512 & 0.656 \\
        S&\checkmark & & & & \checkmark & 0.545 & 0.709 & 0.465 & 0.608 & 0.532 & 0.687 \\
        S& & \checkmark & & & & 0.584 & 0.757 & 0.507 & 0.666 & 0.570 & 0.742\\
        S& & \checkmark & & & \checkmark & 0.592 & 0.772 & 0.518 & 0.686 & 0.581 & 0.758 \\
        S& & & \checkmark & & & 0.600 & 0.791 & 0.519 & 0.698 & 0.586 & 0.766 \\
        S& & & \checkmark & & \checkmark & 0.606 & 0.801 & 0.535 & \textbf{0.722} & 0.588 & 0.777 \\
        S& \checkmark & \checkmark & \checkmark & HW & & 0.614 & 0.794 & 0.532 & 0.692 & 0.602 & 0.776 \\
        S& \checkmark & \checkmark & \checkmark & SM & & 0.612 & 0.787 & 0.539 & 0.697 & 0.607 & 0.777 \\
        S& \checkmark & \checkmark & \checkmark & SW & & 0.615 & 0.808 & 0.534 & 0.705 & 0.600 & 0.780 \\
        S& \checkmark & \checkmark & \checkmark & HW & \checkmark & \textbf{0.627} & \textbf{0.823} & \textbf{0.542} & 0.716 & \textbf{0.606} & \textbf{0.787} \\
        S& \checkmark & \checkmark & \checkmark & SM & \checkmark & 0.591 & 0.761 & 0.501 & 0.649 & 0.575 & 0.736 \\
        S& \checkmark & \checkmark & \checkmark & SW & \checkmark & 0.603 & 0.780 & 0.518 & 0.673 & 0.590 & 0.759 \\
        \hline
    \end{tabular}
    \label{tab:table1}
\end{table*}

\paragraph{Feature Recalibration.} Considering that different channels of deep features play different roles in tracking, we apply SE-blocks \cite{SENet} over the raw deep features extracted with the base feature extractors. An illustration of a SE-block is shown in Fig. \ref{fig:seblock}. The SE-block consists of two steps: 1) \emph{squeeze} and 2) \emph{excitation}. The \emph{squeeze} step corresponds to an average pooling operation. Given a 3D feature map, this operation generates the channel descriptor $\mathbf{\omega}_{sq}$ with
\begin{equation}
    \mathbf{\omega}_{sq}=\frac{1}{W\times H}\sum_{m=1}^{W}\sum_{n=1}^{H}v_{c}(m,n), (c=1,...,C),
\end{equation}
where $W$, $H$, $C$ are the width, height and the number of channels of the deep feature, and $v_{c}(m,n)$ is the corresponding value in the feature map. The subsequent step is the \emph{excitation} through a two-layer Multi-layer perceptron (MLP). Its goal is to capture the channel-wise dependencies that can be expressed as
\begin{equation}
    \mathbf{\omega}_{ex}=\sigma(\mathbf{W_{2}}\delta(\mathbf{W_{1}}\mathbf{\omega}_{sq})),
\end{equation}
where $\sigma$ is a sigmoid activation, $\delta$ is a ReLU activation, $\mathbf{W_{1}}\in \mathbb{R}^{\frac{C}{b}\times C}$ and $\mathbf{W_{2}}\in \mathbb{R}^{C\times\frac{C}{b}}$ are the weights for each layer, and $b$ is the channel reduction factor used to change the dimension. After the \emph{excitation} operation, we obtain the channel weight $\mathbf{\omega}_{ex}$. The weight is used to rescale the feature maps extracted by the base feature extractors with
\begin{equation}
    F_{l_{i}}^{*}=\mathbf{\omega}_{ex}\cdot F_{l_{i}},
\end{equation}
where $\cdot$ is a channel-wise multiplication and $F=(S,A)$. Note that $\mathbf{\omega}_{ex}$ is learned for each layer in a base feature extractor, but the corresponding layers for the CNN branches of the exemplar patch and the search region share the same channel weights. We train the SE-blocks to obtain six $\mathbf{\omega}_{ex}$ in total (see Fig. \ref{fig:architecture}).

\subsection{Response Maps Combination}
Once the recalibrated feature representations from the convolutional layers of each model are obtained, we apply a cross-correlation operation, which is implemented by convolution, over the corresponding feature maps to generate the response map $r$ with
\begin{equation}
    r(z,x)=corr(F^{*}(z),F^{*}(x)),
\end{equation}
where $F^{*}$ is a recalibrated feature map from SiamFC or AlexNet. 

The response maps are then combined. For a pair of image inputs, six response maps are generated, denoted as $r_{c3}^{S}$, $r_{c4}^{S}$, $r_{c5}^{S}$, $r_{c3}^{A}$, $r_{c4}^{A}$ and $r_{c5}^{A}$. Note that we do not need to rescale the response maps for combination, since they have the same size (see Section \ref{sec:detail}, Data Dimensions). The response maps are combined hierarchically. After fusing $r^{S}$ and $r^{A}$ for each of the CNN models, we combine the two resulting response maps to get the final map. The combination is performed by considering three strategies: hard weight (HW), soft mean (SM) and soft weight (SM) \cite{HCFTX}, defined as
\begin{equation}
    \textrm{Hard weight: } r^{*}=\sum_{t=1}^{N}w_{t}r_{t},
\end{equation}
\begin{equation}
    \textrm{Soft mean: } r^{*}=\sum_{t=1}^{N}\frac{r_{t}}{max(r_{t})},
\end{equation}
\begin{equation}
    \textrm{Soft weight: } r^{*}=\sum_{t=1}^{N}\frac{w_{t}r_{t}}{max(r_{t})},
\end{equation}
where $r^{*}$ is the combined response map, $N$ is the number of response maps to be combined together, and $w_{t}$ is an empirical weight for each response map.

The optimal weights $w_{t}$ for HW and SW are obtained experimentally. Finally, the corresponding location of the maximum value in the final response map is the new location of the target.

\section{Experiments}\label{sec:mfst_exp}
The first objective of our experiments is to investigate the contribution of each module in order to find the best response map combination strategy for optimal representations. For this purpose, we perform an ablation analysis. Secondly, we compare our method with the reference SiamFC method and recent state-of-the-art trackers. The experiment results show that our method significantly outperforms SiamFC, while obtaining competitive performance with respect to the recent state-of-the-art trackers.

We performed our experiments on a PC with an Intel i7-3770 3.40 GHz CPU and a Nvidia Titan X GPU. We benchmarked our method on the OTB benchmarks \cite{otb100} and on the VOT2018 benchmark \cite{vot2018}. The benchmark results are calculated using the provided toolkits. The average testing speed of our tracker is 39 fps.

\subsection{Implementation Details}\label{sec:detail}
\paragraph{Network Structure.} We used SiamFC \cite{SiamFC} and AlexNet \cite{AlexNet} as deep feature extractors. The SiamFC network is a fully convolutional neural network, containing five convolutional layers. It has an AlexNet-like architecture, but it is trained on a video dataset for object tracking. The AlexNet network consists of five convolutional layers and three fully connected layers trained on an image classification dataset. We slightly modified the stride of AlexNet to obtain the same dimensions for the outputs of both CNN models. Since only deep features are needed to represent the target, we removed the fully connected layers of AlexNet and only kept the convolutional layers to extract features.

\paragraph{Data Dimensions.} The inputs of our method are the exemplar patch $z$ and the search region $x$. The size of $z$ is $127\times 127$ and the size of $x$ is $255\times 255$. The output feature maps of $z$ have sizes of $10\times 10\times 384$, $8\times 8\times 384$ and $6\times 6\times 256$ respectively. The output feature maps of $x$ have sizes of $26\times 26\times 384$, $24\times 24\times 384$ and $22\times 22\times 256$ respectively. Taking the features of $z$ as filters to perform a convolution on the features of $x$, the size of the output response maps are all the same, $17\times17$. The final response map is resized to the size of the input to locate the target. Since the two feature extractors that we are using are fully convolutional neural networks, the size of inputs can also be adapted to any other dimension.

\begin{table*}[!t]
    \centering
    \caption{Experiments on combining the response maps of the two CNN models. $A_{c5}$ is only taking features from the last convolutional layer of AlexNet network, $S_{c5}$ is only taking features from the last convolutional layer of SiamFC network.  $A_{com}$ is the combined response maps from AlexNet network by soft weight combining, $S_{com}$ is the combined response maps from SiamFC network by hard weight combining. \textbf{Boldface} indicates best results.}
    \begin{tabular}{|c c c c|c c|c c|c c|}
    \hline
        & & & &
      \multicolumn{2}{c|}{OTB-2013} &
      \multicolumn{2}{c|}{OTB-50} &
      \multicolumn{2}{c|}{OTB-100}\\
        $A$ & $S$ & Fusion & SE &  AUC & Prec. & AUC & Prec. & AUC & Prec. \\
    \hline
    $A_{c5}$ & & & &0.582&0.763&0.496&0.665&0.557&0.735\\
    $A_{com}$ & & & & 0.630 & 0.795 & 0.538 & 0.699 & 0.616 & 0.786\\
    $A_{com}$ & & &\checkmark & 0.635 & 0.811 & 0.545 & 0.716 & 0.627 & 0.803\\
    & $S_{c5}$ & & &0.600 & 0.791 & 0.519 & 0.698 & 0.586 & 0.766\\
    & $S_{com}$ & & & 0.614 & 0.794 & 0.532 & 0.692 & 0.602 & 0.776 \\
    & $S_{com}$ & &\checkmark & 0.627 & 0.823 & 0.542 & 0.716 & 0.606 & 0.787\\
    $A_{com}$ & $S_{com}$ & HW & &0.637&0.815&0.555&0.720&0.625&0.801\\
    $A_{com}$ & $S_{com}$ & SM & &0.647&0.819&0.560&0.728&0.638&0.816\\
    $A_{com}$ & $S_{com}$ & SW & &0.647&0.818&0.564&0.734&0.637&0.813\\
    $A_{com}$ & $S_{com}$ & HW & \checkmark&0.667&0.852&\textbf{0.583}&0.761&0.644&0.824\\
    $A_{com}$ & $S_{com}$ & SM & \checkmark & 0.640 & 0.810 & 0.557 & 0.718 & 0.632 & 0.804\\
    $A_{com}$ & $S_{com}$ & SW & \checkmark &\textbf{0.667}&\textbf{0.854}&0.581&\textbf{0.764}&\textbf{0.647}&\textbf{0.831}\\
    \hline
    \end{tabular}
    \label{tab:table2}
\end{table*}

\paragraph{Training.} The SiamFC model is trained on the ImageNet dataset \cite{ImageNet} and only color images are considered. The ImageNet dataset contains more that 4,000 video sequences with about 1.3 million frames and 2 million tracked objects with ground truth bounding boxes. For the input, we take a pair of images and crop the exemplar patch $z$ in the center and the search region $x$ in another image. The SiamFC model is trained with the loss of equation \ref{mainloss} for 50 epochs with an initial learning rate of 0.01. The learning rate decays with a factor of 0.86 after each epoch. The AlexNet model is pretrained on the ImageNet dataset for the image classification task. We just remove the fully connected layers before training the SE-blocks.

After the training of the base feature extractors, we add the SE-blocks in the two models and train them separately in the same manner. For each model, the original parameters are fixed. We then apply SE-blocks on the output of each selected layer (c3, c4 and c5) and take the recalibrated output of each layer as the output feature maps to generate the result for training. The SE-blocks are trained with the videos of the ImageNet dataset with the loss of equation \ref{mainloss} for 50 epochs with an initial learning rate of 0.01. The learning rate decays with a factor of 0.86 after each epoch.

\paragraph{Tracking.} We first initialize our tracker with the initial frame and the coordinates of the bounding box of the target object. After we scaled and cropped the initial frame and obtained the exemplar patch, it is fed into the SiamFC model and AlexNet model to generate the preliminary feature representations $S_{l_{z}}$, $A_{l_{z}} (l=c3, c4, c5)$. Then, the SE-blocks are applied to produce the recalibrated feature maps $S^{*}_{l_{z}}$, $A^{*}_{l_{z}}$, which are then used to produce response maps for tracking the target object for all the following frames.

After the feature maps of the target object are obtained, to track the target, the next frame is fed into the tracker. The tracker crops the region centered on the last center position of the target object, generate the feature representations and output the response maps by a correlation operation with the feature maps of the target object. The corresponding position of the maximum value in the final combined response map indicates the center point of the new position of the target object and the bounding box keeps the same size unless other scales obtain higher response value.

\paragraph{Hyperparameters.} The channel reduction factor $b$ in the SE-blocks is 4. The empirical weights $w_{t}$ for $r_{c3}^{S}$, $r_{c4}^{S}$, $r_{c5}^{S}$, $r_{c3}^{A}$, $r_{c4}^{A}$ and $r_{c5}^{A}$ are 0.1, 0.3, 0.7, 0.1, 0.6 and 0.3. The empirical weights $w_{t}$ for $r^{S}$ and $r^{A}$ are 0.3 and 0.7. To handle scale variations, we search the target object over three scales $1.025^{\{-1, 0, 1\}}$ during evaluation and testing.

\begin{figure*}[t]
    \centering
    \subfigure{\includegraphics[width=0.4\textwidth]{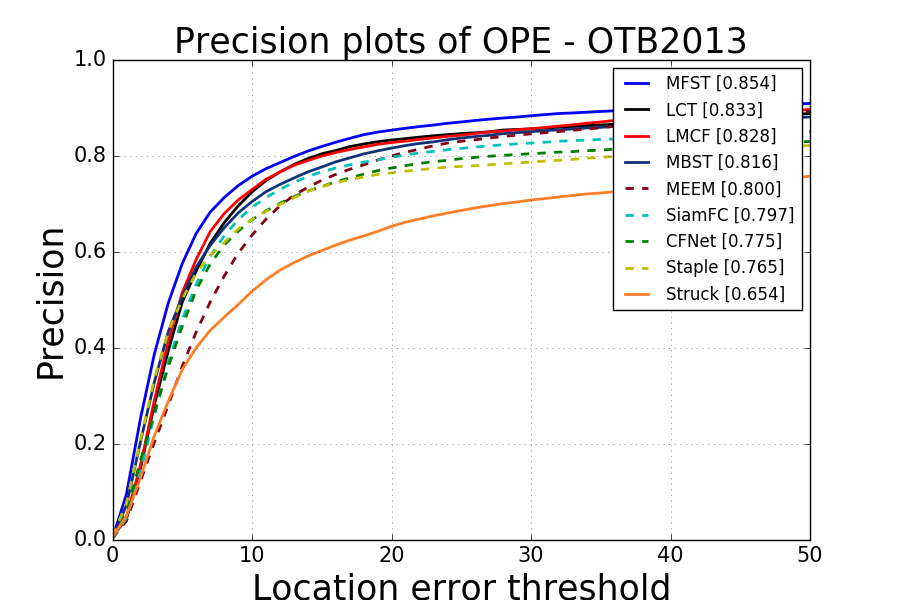}}
    \subfigure{\includegraphics[width=0.4\textwidth]{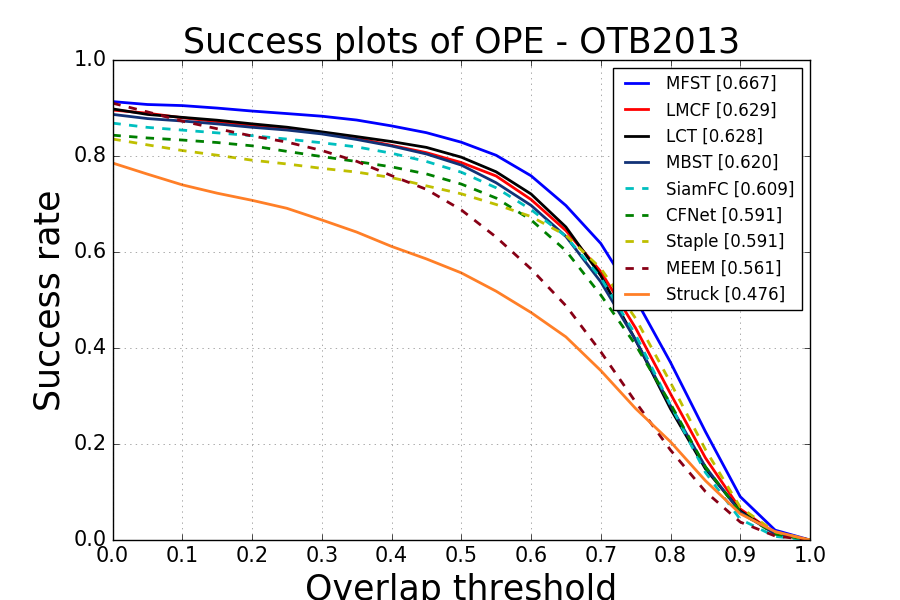}}
    \subfigure{\includegraphics[width=0.4\textwidth]{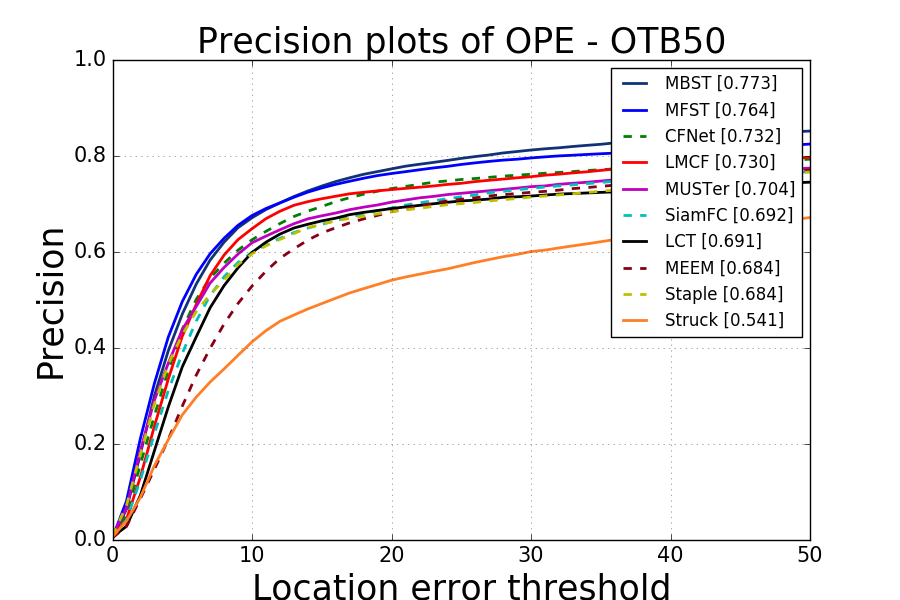}}
    \subfigure{\includegraphics[width=0.4\textwidth]{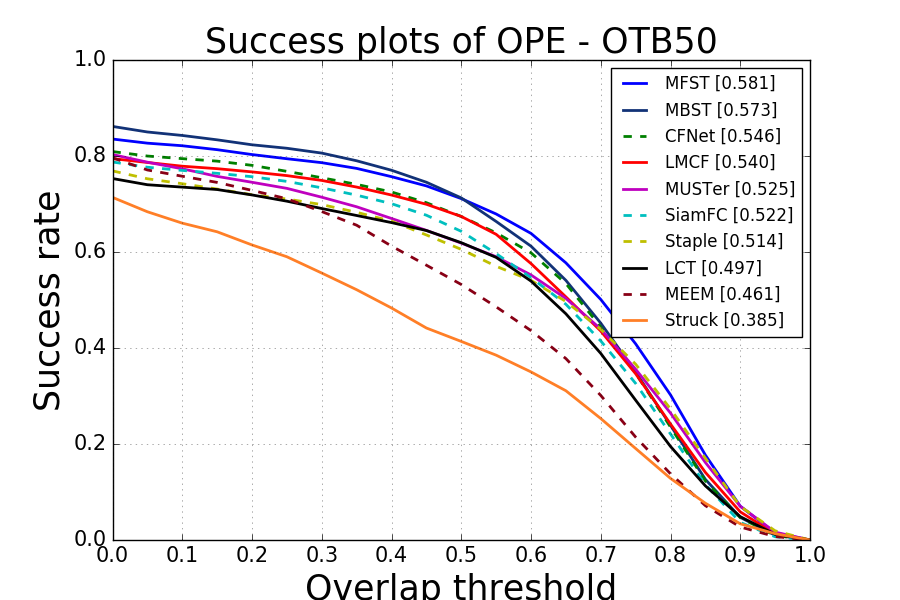}}
    \subfigure{\includegraphics[width=0.4\textwidth]{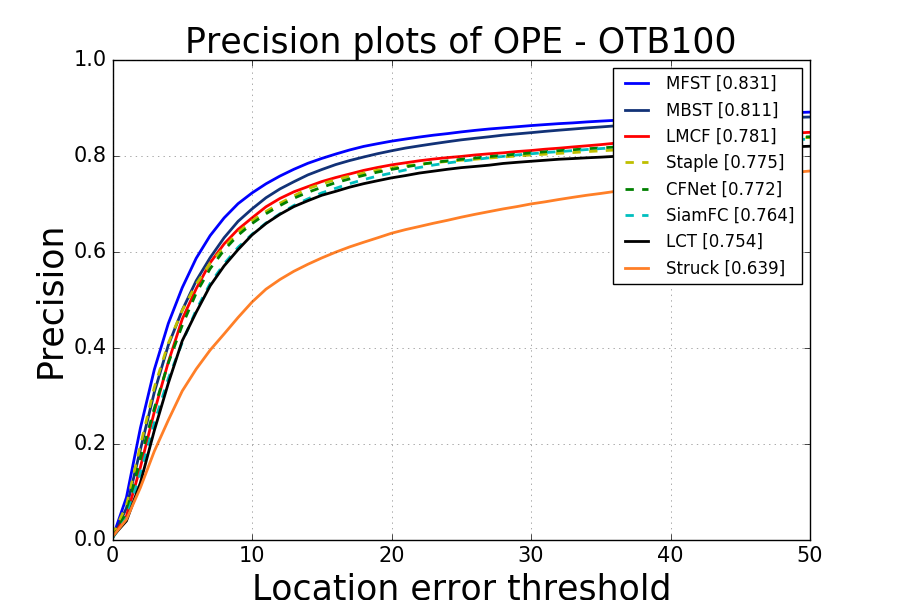}}
    \subfigure{\includegraphics[width=0.4\textwidth]{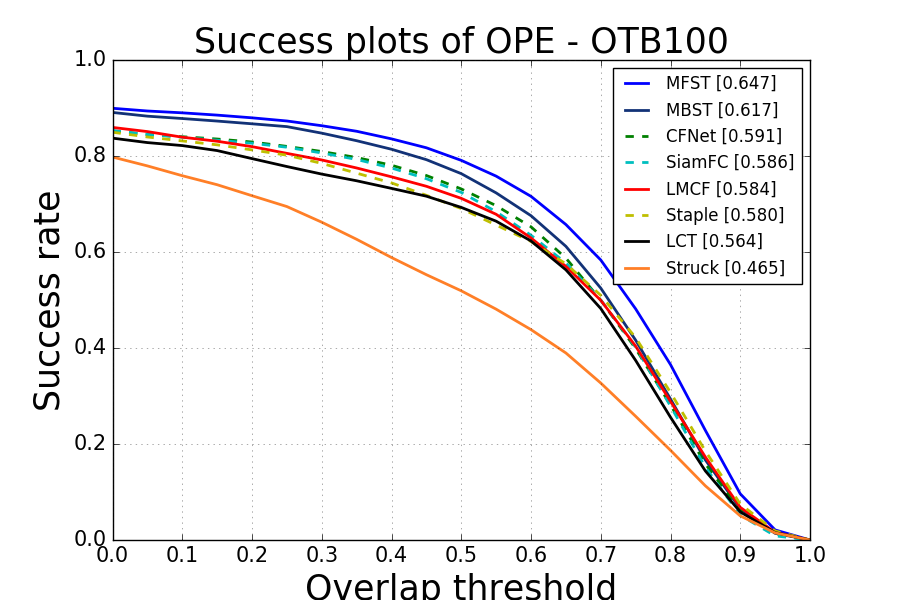}}
    \caption{The evaluation results on OTB benchmarks. The plots are generated by the Python implemented OTB toolkit.}
    \label{fig:cfst_comp}
\end{figure*}

\subsection{Dataset and Evaluation Metrics}
\paragraph{OTB Benchmarks.} We evaluate our method on the OTB benchmarks \cite{otb50, otb100}, which consist of three datasets, OTB50, OTB2013 and OTB100. They contain 50, 51, 100 video sequences with ground truth target labels for object tracking. Two evaluation metrics are used for quantitative analysis, the center location error and the overlap score, which are used to produce precision plots and success plots respectively. To obtain the precision plot, we calculate the the average euclidean distance between the center location of the tracking results and the ground truth labels. The threshold of 20 pixels is used to rank the results. For the success plot, we compute the IoU (intersection over union) between the tracking results and the ground truth labels for each frame. The AUC (area-under-curve) is used to rank the results.

\paragraph{VOT 2018 Benchmark.} VOT2018 short-term benchmark consists of 60 video sequences, the target in the sequences are annotated by a rotated bounding-box. The benchmark takes three primary measures to evaluate the tracking performance: accuracy (A), robustness (R) and expected average overlap (EAO). The accuracy is calculated by the average overlap between the tracker predictions and the ground truth bounding boxes, while the robustness is how many times the target get lost during tracking. The last evaluation metric, expected average overlap, measures the expected average overlap of a tracker when given sequences with the same visual properties. The VOT benchmark utilizes a reset-based methodology, which means that the tracker is re-initialized when its prediction has no overlap with the ground truth. 

\subsection{Ablation Study}
To investigate the contributions of each module and the optimal strategies to combine representations, we performed an ablation study with several variations of our method. We first studied the combination strategy that achieves the best performance on the OTB benchmarks to generate the combined response maps for each model, which are denoted as $r^{S}$ and $r^{A}$ (see Table \ref{tab:table1}). After that, as illustrated in Table \ref{tab:table2}, we test the three different strategies again to find the best strategy to combine $r^{S}$ and $r^{A}$.

\paragraph{A proper combination of features is better than features from single layer.} As illustrated in Table \ref{tab:table1}, we experimented using features from a single layer as the target representation and combined features from several layers with different combination strategies for the two CNN models. The results show that, taken separately, c3, c4, c5 give results that are approximately similar. Since object appearance changes, c3 that should give the most precise location does not always achieve good performance. However, with a proper combination, the representation power of the combined feature gets much improved.

\paragraph{Features get enhanced with recalibration.} Due to the \emph{squeeze} and \emph{excitation} operations, recalibrated features achieves better performance than the preliminary features. Recalibration through SE-blocks thus improves the representation power of features from single layer, which results in a better representation of the combined features.

\paragraph{Multiple models are better than a single model.} Our approach utilizes two CNN models as feature extractors. Therefore, we also conducted experiments to verify the benefit of using two CNN models. As illustrated in Table \ref{tab:table2}, we evaluated the performance of using each CNN model separately and using the combination of two CNN models. The results show that the combination of two models is more discriminative than only one model regardless of the use SE-blocks.

\paragraph{A proper strategy is important for the response map combination.} We applied three strategies to combine the response maps: hard weight (HW), soft mean (SM) and soft weight (SW). Since the two CNN models are trained for different tasks and features from different layers embed different level of information, different types of combination strategies should be applied to make the best use of the features. The experimental results show that generally, combined features are more discriminative than independent features, while a proper strategy can improve the performance significantly as illustrated in Table \ref{tab:table1} and Table \ref{tab:table2}. In addition, we observe that the soft weight strategy is generally the most appropriate, except for combining hierarchical features from the SiamFC model.

\begin{table*}[!t]
    \centering
    \caption{The speed evaluation results on OTB benchmarks.}
    \begin{tabular}{|c|c|c|c|c|c|c|c|c|c|c|}
    \hline
        \textbf{Tracker} & MFST & MBST & LCT & LMCF & MEEM & CFNet & SiamFC & Staple & Struck & MUSTer \\
        \hline
        
        \textbf{Speed(fps)} & 39 & 17 & 27 & 66 & 22 & 75 & 86 & 41 & 10 & 5
        \\
    \hline
    \end{tabular}
    \label{tab:speed_table}
\end{table*}

\begin{table*}[!t]
    \centering
    \caption{Evaluation results of our trackers and some recent the state-of-the-art trackers on VOT2018 benchmark. \textcolor{blue}{Blue}: best, \textcolor{orange}{Orange}: second best, \textcolor{red}{Red}: third best. $\uparrow$: higher is better, $\downarrow$: lower is better. A: Accuracy, R: Robustness, AO: Average overlap, EAO: Expected AO. For the unsupervised experiment, the tracker is not re-initialized when it fails. For the real-time experiment, frames are skipped if the tracker is not fast enough.}
     \begin{tabular}[width=0.6\linewidth]{|c|c|c|c|c|c|c|c|c|c|}
    \hline
    \textbf{} & \multicolumn{3}{c|}{\textbf{baseline}} & \multicolumn{3}{c|}{\textbf{real-time}} & \multicolumn{2}{c|}{\textbf{unsupervised}} \\
    \hline
    \textbf{Tracker} & \textbf{EAO}$\uparrow$ & \textbf{A}$\uparrow$ & \textbf{R}$\downarrow$ & \textbf{EAO}$\uparrow$ & \textbf{A}$\uparrow$ & \textbf{R}$\downarrow$  & \textbf{AO}$\uparrow$ & \textbf{Speed (FPS)}$\uparrow$  \\
    \hline
    MEEM & 0.192 & 0.463 & \textcolor{red}{0.534} & 0.072 & 0.407 & 1.592 & 0.328 & 4.9 \\
    \hline
    KCF & 0.094 & 0.417 & 1.726 & 0.088 & 0.428 & 1.926 & 0.174 & N/A \\
    \hline
    Staple & 0.169 & \textcolor{orange}{0.530} & 0.688 & 0.170 & \textcolor{orange}{0.530} & 0.688 & 0.335 & 54.3\\
    \hline
    ANT & 0.168 & 0.464 & 0.632 & 0.059 & 0.403 & 1.737 & 0.279 & 4.1 \\
    \hline
    CFNet & \textcolor{orange}{0.188} & 0.503 & 0.585 & \textcolor{red}{0.182} & 0.502 & \textcolor{red}{0.604} & \textcolor{red}{0.345} & 42.6 \\
    \hline
    SiamFCOSP & 0.171 & \textcolor{red}{0.508} & 1.194 & 0.166 & \textcolor{red}{0.503} & 1.254 & 0.241 & N/A \\
    \hline
    ALTO & 0.182 & 0.358 & 0.818 & 0.172 & 0.365 & 0.888 & 0.252 & 42.9 \\
    \hline
    SiamRPN++ & \textcolor{blue}{0.285} & \textcolor{blue}{0.599} & \textcolor{orange}{0.482}  & \textcolor{blue}{0.285} & \textcolor{blue}{0.599} & \textcolor{orange}{0.482} & \textcolor{blue}{0.482} & 36.3\\\hline\hline
    \textbf{MFST} & \textcolor{red}{0.200} & 0.497 & \textcolor{blue}{0.428} & \textcolor{orange}{0.200} & 0.488 & \textcolor{blue}{0.455} & \textcolor{orange}{0.348} & 33.2\\\hline
    \end{tabular}
    \label{tab:vot18_all}
\end{table*}

\subsection{Comparisons}
We compare our tracker MFST with MBST \cite{MBST}, LMCF \cite{LMCF}, CFNet \cite{CFNet}, SiamFC \cite{SiamFC},  Staple \cite{Staple}, Struck \cite{Struck}, MUSTER \cite{MUSTer}, LCT \cite{LCT}, MEEM \cite{MEEM} on OTB benchmarks \cite{otb50,otb100}. The precision plot and success plot are shown in Fig. \ref{fig:cfst_comp}. Both plots show that our tracker MFST achieves the best performance among these recent state-of-the-art trackers on OTB benchmarks, except on the OTB-50 benchmark precision plot. It demonstrates that by using the combined features, the target representation of our method is more robust then our base tracker SiamFC. The feature calibration mechanism we employed is beneficial for tracking as well. Although we use siamese networks to address the tracking problem as for SiamFC, and take SiamFC as one of our feature extractor, our tracker achieves much improved performance over SiamFC. Besides, despite the fact that MBST tracker employs diverse feature representations from many CNN models, our tracker achieves better results with only two CNN models, in terms of both tracking accuracy and speed. 

A speed comparison is shown in Table \ref{tab:speed_table} and Table \ref{tab:vot18_all} (Speed). Because we use two feature extractor networks, our MFST is slower then SiamFC. Still, it is faster than several trackers in the literature that are less robust. Our method shows a better speed vs accuracy compromise than MBST that combines features from several base feature extractor networks.


In addition to the OTB benchmarks\cite{otb50, otb100}, we evaluate our MFST tracker on the VOT2018 benchmark\cite{VOT_TPAMI, vot2018} and compared it with some recent and classic state-of-the-art trackers, including MEEM \cite{MEEM}, some correlation-based trackers: KCF \cite{KCF}, Staple\cite{Staple}, ANT\cite{ANT}, and several Siamese-based trackers: CFNet\cite{CFNet}, SiamFCOSP \cite{VOT2019}, ALTO \cite{VOT2019} and SiamRPN++ \cite{Li2019SiamRPNEO}. The results are produced by the VOT toolkit\cite{VOT_TPAMI} and reported in Table \ref{tab:vot18_all}. These results show that our method is more robust to the compared trackers with less failures as depicted by the R value. On that aspect, our tracker does better than SiamRPN++, demonstrating that our feature and fusion approach helps in better representing the target. However, it seems that using proposal, like in SiamRPN++ can lead to better accuracy (higher A value). Proposals could be included in our method. Our method ranks a little better for EAO compared to A for the baseline and real-time scenarios. This shows that our features generalize better then the one used by other trackers. Moreover, it is interesting to note that our tracker also performs well in the unsupervised scenario where the tracker is not reset after failure, showing the robustness of our siamese tracker compared to CFNet, SiamFCOSP and ALTO, which, like our tracker, do not use a region proposal network. Although our tracker is not the fastest siamese tracker, it is fast enough to maintain good performance in the real-time scenario, where frames are skipped if the tracker is not fast enough to process a video at 20 FPS.

\section{Conclusion}\label{sec:mfst_conclude}
In this paper, we presented a Multiple Features-Siamese Tracker (MFST) that exploits diverse features at different convolutional layers within the Siamese tracking framework. We utilize features from different hierarchical levels and from different models using three combination strategies. Based on the feature combination, different levels of abstraction of the target are encoded into a fused feature representation. Moreover, the tracker greatly benefits from the new feature representation due to a calibration mechanism applied to different channels to recalibrate features. As a result, MFST achieved strong performance with respect to recent state-of-the-art trackers on object tracking benchmarks.

\section*{Acknowledgments}
This work was supported by The Fonds de recherche du Québec - Nature et technologies (FRQNT) and Mitacs. We also thank Nvidia for providing us with the Nvidia TITAN X GPU.

\end{document}